\documentclass[11pt,a4paper]{article}
\usepackage[hyperref]{acl2019}
\usepackage{times}
\usepackage{latexsym}
\usepackage{bm}
\usepackage{amsmath}
\usepackage{url}
\usepackage{makecell}
\usepackage{multirow}
\usepackage{subfigure}
\usepackage{graphicx} 
\usepackage{ulem}
\usepackage{colortbl}
\definecolor{mypink}{rgb}{.99,.91,.95}
\aclfinalcopy
\def\aclpaperid{5} 

\title{The University of Sydney's Machine Translation System for WMT19}

\author{%
  Liang Ding\quad \quad Dacheng Tao\\
  UBTECH Sydney AI Center, School of Computer Science, FEIT\\
  University of Sydney, Australia\\
  {\tt ldin3097@uni.sydney.edu.au, dacheng.tao@sydney.edu.au}
 }

\date{}

\begin{document}
\maketitle
\begin{abstract}
  This paper describes the University of Sydney's submission of the WMT 2019 shared news translation task. 
  We participated in the Finnish$\rightarrow$English direction and got the best BLEU(33.0) score among all the participants. 
  Our system is based on the self-attentional Transformer networks, into which we integrated the most recent effective strategies from academic research (\textit{e.g.}, BPE, back translation, multi-features data selection, data augmentation, greedy model ensemble, reranking, ConMBR system combination, and post-processing). Furthermore, we propose a novel augmentation method $\textbf{Cycle Translation}$ and a data mixture strategy $\bm{Big}$/$\bm{Small}$ \textbf{ parallel construction} to entirely exploit the synthetic corpus. Extensive experiments show that adding the above techniques can make continuous improvements of the BLEU scores, and the best result outperforms the baseline (Transformer ensemble model trained with the original parallel corpus) by approximately 5.3 BLEU score, achieving the state-of-the-art performance.
\end{abstract}

\section{Introduction}
  Neural machine translation (NMT), as a succinct end-to-end paradigm, has resulted in massive leap in state-of-the-art performances for many language pairs~\cite{D13-1176,seq2seq,rnnsearch,convs2s,gnmt,transformer}. Among these encoder-decoder networks, the Transformer~\cite{transformer}, which solely uses along attention mechanism and eschews the recurrent or convolutional networks, leads to state-of-the-art translation quality and fast convergence speed~\cite{ahmed2017weighted}.
  Although many Transformer-based variants are proposed (\textit{e.g.}, DynamicConv~\cite{wu2018pay}, sparse-transformer~\cite{child2019generating}), our preliminary experiments show that their performances are unstable compared to the traditional Transformer. Traditional Transformer therefore was employed as our baseline system. In this paper, we summarize the USYD NMT systems for the WMT 2019 Finnish$\rightarrow$English (FI$\rightarrow$EN) translation task.
 
    \begin{table}[t!]
    \begin{center}
    \begin{tabular}{|l|p{6.4cm}|}
    \hline 
    \# & \textbf{cycle translated sample sentence pair}\\
    \hline
    1 & \textit{She stuck to her principles even when some suggest that in an environment often considered devoid of such thing there are little point.}\\ 
    \hline
    2 & \textit{She insists on her own principles, even if some people think that it doesn't make sense in an environment that is often considered to be absent.}\\ 
    \hline
    \end{tabular}
    \end{center}
    \caption{\label{tab:sentences}Example of difference between original sentence (line 1) and cycle translated result (line 2). Pre-trained BERT model using all available English corpora show that the $\mathcal{L}oss$ decreased from 6.98 to 1.52.}
    \end{table}
  
\begin{figure*}[htb]
    \centering
    \includegraphics[width=0.98\textwidth]{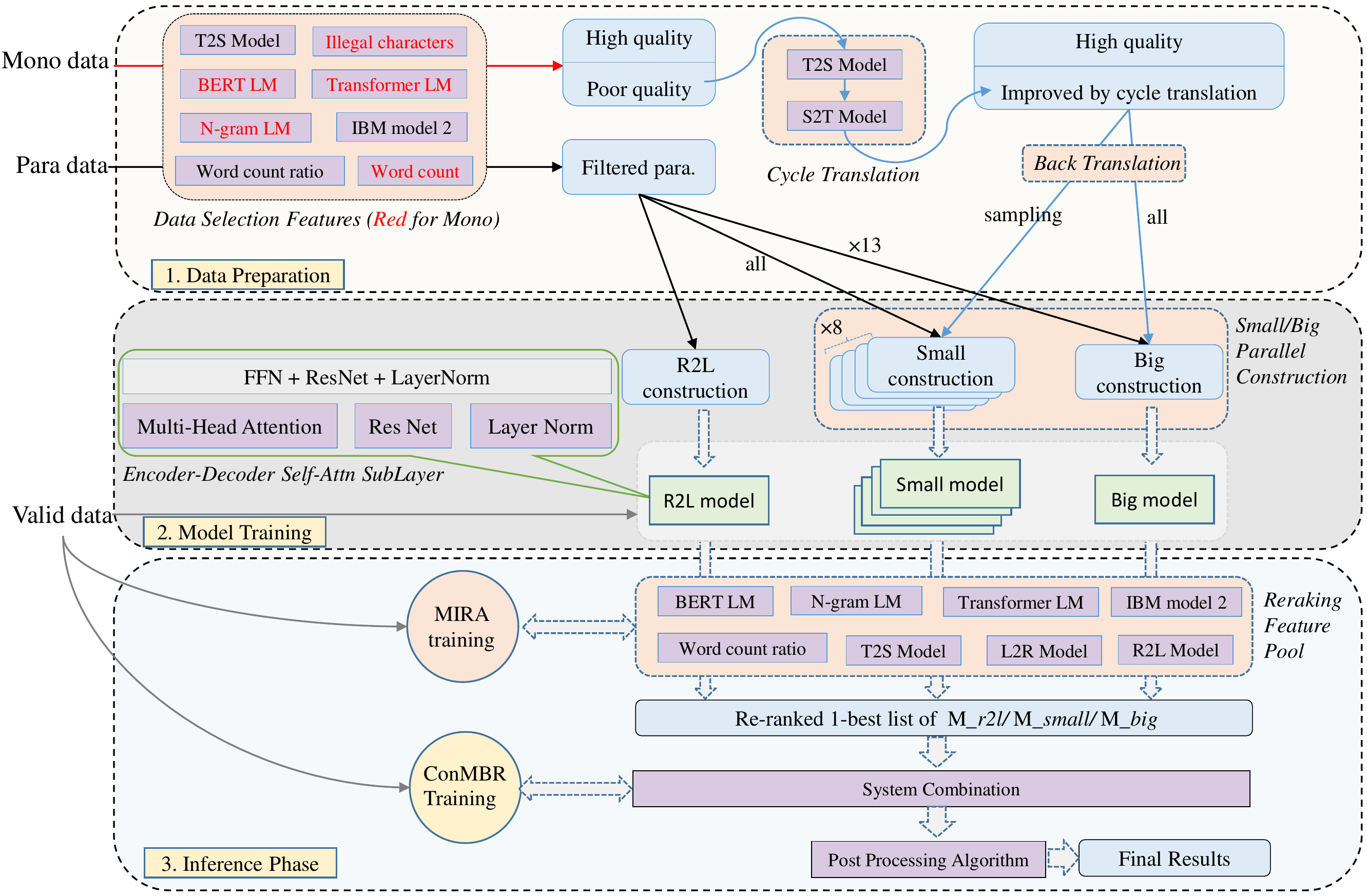}
    \caption{The schematic structure of the three main stages of the USYD-NMT. They are data preparation stage, model training stage and inference phrase. For brevity, here Mono, Para, and Valid represent the monolingual, parallel and validation data respectively.}
    \label{fig:system}
\end{figure*}
  
  As the limitation of time and computation resources, we only participated in one challenging task FI$\rightarrow$EN, which lags behind other language pairs in translation performance~\cite{bojar-etal-2018-findings}. We introduce our system with three parts. 
  
  First, at data level, we find that the data quality of both parallel and monolingual is unbalanced (\textit{i.e.}, contains a large number of low quality sentences). Thus, we apply several features to select the data after pre-processing, for example, language models, alignment scores etc. Meanwhile, in order to fully utilize monolingual corpus, not only back translation~\cite{sennrich2015improving} is adopted to back translate the high quality monolingual sentences with target-to-source(T2S) model, we also propose $\textbf{Cycle Translation}$ to improve the low-quality sentences, in turn resulting in corresponding high-quality back translation results. Note that unlike text style transfer task~\cite{shen2017style,fu2018style,prabhumoye2018style} which transfers text to specific style (\textit{e.g.}, political slant, gender), we aim to improve the fluency of sentences, for instance, through cycle translation, low quality sentence in Table~\ref{tab:sentences} becomes more fluent in terms of language model score. The top diagram of Figure~\ref{fig:system} depicts data preparation process concretely.
  
  As to model training in the middle part of Figure~\ref{fig:system}, we empirically introduced $\textbf{Big/Small parallel construction}$ strategy to construct training data for different models. The intuition is all the data are advantageous and can be fully exploited by different models, thus we train $8$ Transformer\_base models ($\mathcal{M}_{small}\times8$) by using different small scale corpus constructed by small parallel construction method and a Transformer\_big model ($\mathcal{M}_{big}\times1$) based on the big parallel construction method. In the meantime, a right-to-left model ($\mathcal{M}_{r2l}$) is trained.
  
  In addition, in inference phrase, we comprehensively consider the ensemble strategies at model level, sentence level and word level. For model level ensemble, while brutal ensemble top-$N$ or last-$M$ models may improve translation performance, it is difficult to obtain the optimal result. Hence we employ Greedy Model Selection based Ensembling (GMSE)~\cite{partalas2008focused,deng-etal-2018-alibabas}. For sentence level ensemble, we keep top n-best for multi-features reranking. And for word aspect, we adopt the confusion network decoding~\cite{bangalore2001computing,matusov2006computing,sim2007consensus} with using the consensus network minimum Bayes risk (MBR) criterion~\cite{sim2007consensus}. After combination, a post-processing algorithm is employed to correct inconsistent number and years between the source and target sentences. The bottom part of Figure~\ref{fig:system} shows the inference process.
  
  Our omnivorous model achieved the best BLEU~\cite{papineni2002bleu} scores among submitted systems, demonstrating the effectiveness of the proposed approach. Theoretically, our approach is not specific to the Finnish$\rightarrow$English language pair, \textit{i.e.}, it is universal and effective for any language pairs. The remainder of this article is organized as follows: Section~\ref{sec:app} will describe each component of the system. In Section~\ref{sec:data}, we introduce the data preparing details. Then, the experimental results are showed in Section~\ref{sec:exp}. Finally, we conclude in Section~\ref{sec:con}.

\begin{table}[t!]
    \begin{center}
    \begin{tabular}{c|c|c}
    \hline 
    model\_parameters & $\mathcal{M}$\_{small} & $\mathcal{M}$\_{big} \\
    \hline
    num\_stack & 6 & 6\\ 
    \hline
    hidden\_size & 512 & 1024\\
    \hline
    FFN\_size & 2048 & 4096\\
    \hline
    num\_heads & 8 & 16\\
    \hline
    p\_dropout & 0.1 & 0.3\\
    \hline
    \end{tabular}
    \end{center}
    \caption{\label{tab:models}Model differences between base and big.}
\end{table}

\section{Approach}
\label{sec:app}

\subsection{Neural Machine Translation Models}
\label{ssec:nmt}

Given a source sentence $X={x_1,...,x_{T'}}$, NMT model factors the distribution over target sentence $Y={y_1,...,y_T}$ into a conditional probabilities:
\begin{equation}
    p(Y|X;\theta)=\prod_{t=1}^{T+1} p(y_t|y_{0:t-1},x_{1:T'};\theta)
\end{equation}
where the conditional probabilities are parameterized by neural networks. 

The NMT model consists of two units: an encoder and a decoder. The encoder is assumed that it can adequately represent the source sentence. Then, the decoder can recursively predict each target word. Parameters of encoder, decoder and attention mechanism are trained to maximize the likelihood with a cross-entropy loss applied:

\begin{equation}
\begin{aligned}
    \mathcal{L}_{ML} &= \log p(Y|X;\theta)
    \\
    {} &= \sum_{t=1}^{T+1}\log p(y_t|y_{0:t-1},x_{1:T'};\theta)
\end{aligned}
\end{equation}

Concretely, an self-attentional encoder-decoder architecture~\cite{transformer} was selected to capture the causal structure. For training with different size of corpus, we employ the Transformer\_base ($\mathcal{M}$\_{\textbf{base}}) and Transformer\_big ($\mathcal{M}$\_{\textbf{big}}) in our structure, see Table~\ref{tab:models}.

\subsection{Data Selection Features}
\label{ssec:data}

Inspired by~\cite{bei-etal-2018-empirical}, where their system shows data selection can obtain substantial gains, we deliberately design criteria for parallel and monolingual corpus. Both of them employ rule-based features, count features, language model features. And for parallel data, word alignment-based features, T2S translation model score features are applied. The feature types are described in Table~\ref{tab:data-features}. Our BERT language model used here is trained from scratch by the open-source tool\footnote{\url{https://github.com/huggingface/pytorch-pretrained-BERT}} with target side data.

According to our observations, by using above multiple data selection filters, issues like misalignment, translation error, illegal characters, over translation and under translation in terms of length could be significantly reduced.

\begin{table}[t!]
    \begin{center}
    \begin{tabular}{|c|l|}
    \hline 
    \small{Category} & \small{Features}\\
    \hline
    \small{NMT Features} & \small{T2S score}~\cite{sennrich-etal-2016-edinburgh}\\ 
    \hline
    \multirowcell{3}{\small{LM Features}} & \small{BERT LM}~\cite{devlin2018bert}\\
    \cline{2-2}
     & \small{Transformer LM}~\cite{bei-etal-2018-empirical}\\
     \cline{2-2}
     ~ & \small{N-gram LM}~\cite{stolcke2002srilm} \\
     \hline
     \small{Alignment Features} & \small{IBM model 2}~\cite{dyer-etal-2013-simple}\\
     \hline
     \small{Rule-based features} & \small{Illegal characters~\cite{bei-etal-2018-empirical}}\\
     \hline
     \multirowcell{2}{\small{Count Features}} & \small{Word count}\\
     \cline{2-2}
      ~& \small{Word count ratio}\\
     \hline
    \end{tabular}
    \end{center}
    \caption{\label{tab:data-features} Features for data selection.}
\end{table}

\subsection{Cycle Translation for Low-quality Data}

\begin{figure}[ht]
    \centering
    \includegraphics[width=0.48\textwidth]{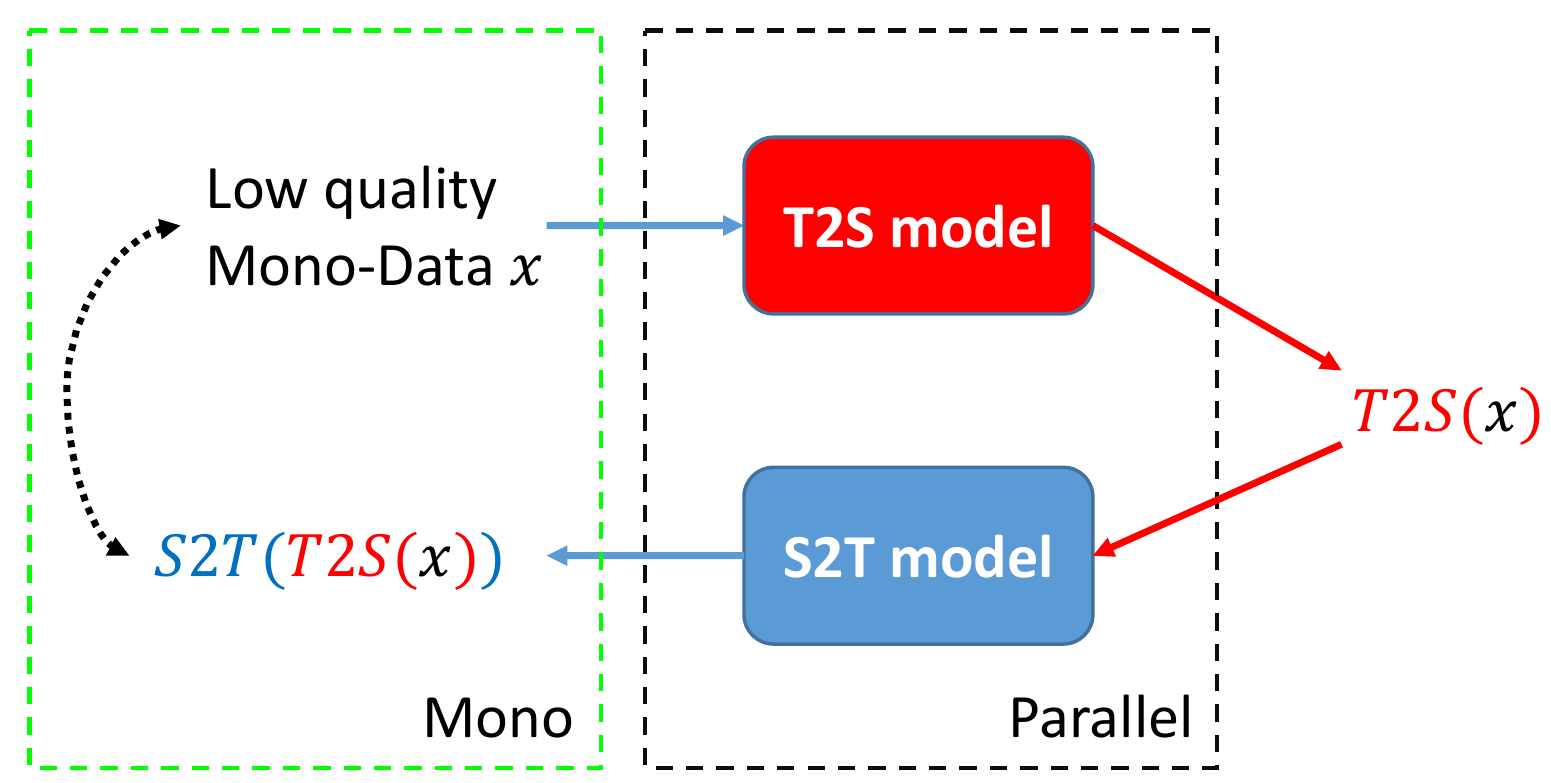}
    \caption{The Cycle Translation process, into which we feed the low quality monolingual data $x$, and then correspondingly obtain the improved data $\mathcal{C}\mathcal{T}(x)$ (denoted as $S2T(T2S(x))$ in figure). Note that models marked in red and green represent the T2S and S2T model trained by $\mathcal{M}_{small}$ with the processed given parallel corpus, the red arrows indicate the data flows of the opposite language type of the inputs. The dotted double-headed arrow between the input $x$ and the final output $\mathcal{C}\mathcal{T}(x)$ means that they share the semantics but differs in fluency.}
    \label{fig:cycle-translation}
\end{figure}

Although the data selection procedure has preserved relatively high quality monolingual data, there are still a large batch of data is incomplete or grammatically incorrect. To address this problem, we proposed Cycle Translation (denoted as $\mathcal{C}\mathcal{T}(\cdot)$, as Figure~\ref{fig:cycle-translation}) to improve the mono-lingual data that below the quality-threshold (According to our empirical ablation study in section~\ref{sec:exp}, the latter 50\% will be cycle translated in our submitted system).

\begin{figure}[ht]
    \centering
    \includegraphics[width=0.47\textwidth]{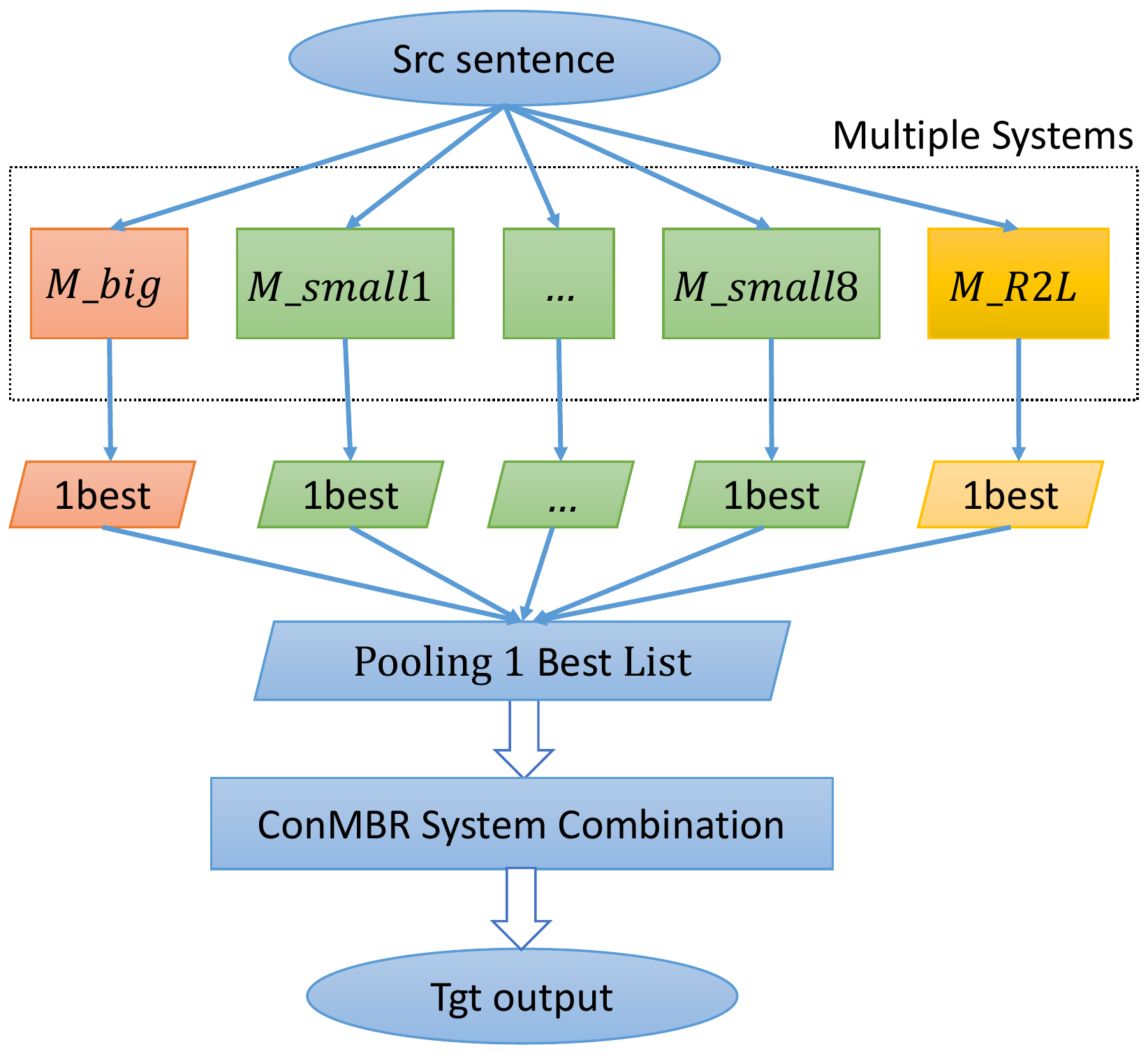}
    \caption{The System Combination process, into which we feed each system/model with the source sentence $x$, in turn obtain corresponding 1-best result $\mathcal{M}_{big}(x)$, $\mathcal{M}_{small1}(x)$, ... ,$\mathcal{M}_{small2}(x)$,$\mathcal{M}_{R2L}(x)$ (Note that the 1-best result here of each system was already reranked). After pooling all system results, we can perform the ConMBR system combination decoding and obtain the final target side results.}
    \label{fig:system-comb}
\end{figure}

\subsection{Back Translation for monolingual corpus}

Back-translation~\cite{sennrich2015improving,bojar-etal-2018-findings}, translating the large scale monolingual corpus to generate synthetic parallel data by Target-to-Source pretrained model, has been widely utilized to improve the translation quality since adding the synthetic data into parallel data can enhance the in-domain information over the original corpus distributions, allowing the translation model to be more robust and deterministic.

\subsection{Greedy Model Selection Based Ensemble}
\label{ssec:GMSE}
Model ensemble is a typical boosting technique, which refers to combining multiple models to reduce stochastic differences in the output that may not be avoided at a single run. Also normally, ensemble model outperforms the the best single one. In neural machine translation, we generally ensemble several checkpoints saved during a single model training. However, our preliminary experiments show that both top-N or last-M ensembling approaches could only bring very insignificant improvements but consume a lot of GPU resources.

To overcome this issue, we adopt greedy model selection based ensembling(GMSE), which technically follows the instruction of~\cite{deng-etal-2018-alibabas}.

\subsection{Reranking n-best Hypotheses}
\label{ssec:rerank}

As the NMT decoding being generally from left to right, this leads to label bias problem~\cite{lafferty2001conditional}. To alleviate this problem, we rerank the n-best hypotheses through training a $k$-best batch MIRA ranker~\cite{Cherry:2012:BTS:2382029.2382089} with multiple features on validation set. The feature pool we integrated include left-to-right (L2R) translation model, (right-to-left) R2L translation model, (target-to-source) T2S translation model, language model, IBM model 2 alignment score, and word count ratio. After multi-feature reranking, the best hypothesis of each model ($\mathcal{M}_{big}\times1$, $\mathcal{M}_{small}\times8$ and R2L model) was retained for system combination.

\subsubsection{Left-to-right NMT model}
The L2R feature refers to the original translation model that could generate the $n$-best list. During reranking training, we keep the original perplexity score evaluated by this L2R model as L2R feature.

\subsubsection{Right-to-Left NMT Model}
The R2L NMT model using the same training data but with inverted target sentences (\textit{i.e.}, reverse target side characters ``a b c d''$\rightarrow$``d c b a''). Then, inverting the hypothesis in the $n$-best list such that each sequence can be given a perplexity score by R2L model.

\subsubsection{Target-to-Source NMT Model}
The T2S model was initially trained for back-translation, we can employ this model to assess the translation adequacy as well by adding the T2S feature to reranking feature pool.

\subsubsection{Language Model}
Besides above features, we employ language models as an auxiliary feature to give the fluent sentences better scores such that the results are easier to understand by human.

\subsubsection{Word Count Ratio}
To alleviate over-translation or under-translation in terms of length, we set the optimal ratio of $\mathcal{L}_{fi}:\mathcal{L}_{en}$ to 0.76 according to the corpus-based statistics. We use the deviation between the ratio of each sentence pair and this optimal ratio as the score.

\subsection{System Combination}
\label{ssec:combination}
 As is shown in Figure~\ref{fig:system-comb}, in order to take full advantages of different models($\mathcal{M}_{big}\times1$, $\mathcal{M}_{small}\times8$ and R2L model), we adopted word-level combination where confusion network was built. Concretely, our method follows Consensus Network Minimum Bayes Risk (ConMBR)~\cite{sim2007consensus}, which can be modeled as
\begin{equation}
    E_{ConMBR}=\mathrm{argmin}_{E'} \mathcal{L}(E',E_{con})
\end{equation}
where $E_{con}$ was obtained as backbone through performing consensus network decoding.

\begin{table}[t!]
    \begin{center}
    \begin{tabular}{l|p{5.74cm}}
    \hline
    src & \textit{Siltalan edellinen kausi liigassa oli \uwave{2006-07}}\\ 
    \hline
    pred & \textit{Siltala's previous season in the league was \uwave{2006 at 07}}\\ 
    \hline
    +post & \textit{Siltala's previous season in the league was \uwave{2006-07}}\\ 
    \hline
    \end{tabular}
    \end{center}
    \caption{\label{tab:post-process}Example of the effectiveness of post-processing in handling inconsistent number translation.}
\end{table}

\subsection{Post-processing}
\label{ssec:post}

In addition to general post-processing strategies (\textit{i.e.}, de-BPE, de-tokenization and de-truecase~\footnote{\url{https://github.com/moses-smt/mosesdecoder/tree/master/scripts}}), we also employed a post-processing algorithm~\cite{wang-etal-2018-niutrans} for inconsistent number, date translation, for example, ``\textit{2006-07}'' might be segmented as ``\textit{2006 -@@ 07}'' by BPE, resulting in the wrong translation ``\textit{2006 at 07}''. Our post-processing algorithm will search for the best matching number string from the source sentence to replace these types of errors, see Table~\ref{tab:post-process}.

\begin{table}
    \begin{center}
    \begin{tabular}{|l|l|}
    \hline
    \textbf{Data} & \textbf{Sentences}\\ 
    \hline
    filtered parallel corpus& 5,831,606\\ 
    reconstructed mono & 82,773,126\\
    filtered synthetic parallel & 75,940,978\\
    \hline
    small construction(\small{$\times8$}) & 11,663,212\\
    big construction & 151,751,856\\
    \hline
    \end{tabular}
    \end{center}
    \caption{\label{tab:statics}Data statistics after data preparation}
\end{table}

\section{Data Preparation}
\label{sec:data}
We used all available parallel corpus~\footnote{both parallel and monolingual corpus can be obtained from: \url{http://www.statmt.org/wmt19/translation-task.html}} for Finnish$\rightarrow$English except the ``Wiki Headlines'' due to the large number of incomplete sentences, and for monolingual target side English data, we selected all besides the ``Common Crawl'' and ``News Discussions''. The criteria is inspired by~\cite{marie-etal-2018-nicts}, who won the first place in this direction at WMT18. Table~\ref{tab:statics} shows the final corpus statistics. More details are as follows:\\

\textbf{Parallel Data}:
We use the criteria in section~\ref{ssec:data}, the overall criteria are following:

\begin{itemize}
\item Remove duplicate sentence pairs.
\item Remove sentence pairs containing illegal characters.
\item Retain sentence pairs between 3 and 80 in length.
\item Remove sentence pairs that are too far from the best ratio($\mathcal{L}_{fi}:\mathcal{L}_{en}$=0.76)
\item Remove pairs containing influent English sentences according to a series of LM features.
\item Remove inadequate translation sentence pairs according to $\mathcal{M}_{T2S}$ score.
\item Remove sentence pairs with poor alignment quality according to IBM model 2.
\end{itemize}

After data selection, there are approximately 5.8M parallel sentences.\\

\begin{table*}[hbt]
    \begin{center}
    \begin{tabular}{l|l|c|c|c}
    \hline
    \textbf{\#} & \textbf{Models} & news-test18 & news-test19 & $\Delta_{ave}$\\ 
    \hline
    $1$ & Baseline\small{(original\_parallel + ensemble)} & 21.8 & 27.3 & $-$ \\
    \hline
    $2$ & $\mathcal{M}_{small}$\small{(selected\_parallel)} & 22.6 & 27.9 & $+0.70$ \\
    \hline
    $3$ & \verb|+synthetic| & 23.9 & 28.8 & \\
    $4$ & \verb|+GMSE| & 24.2 & 29.2 & \\
    $5$ & \verb|+reranking| & 24.6 & 29.5 & \\
    $6$ & \verb|+post processing| & 24.8 & 29.6 & $+2.65$\\
    \hline
    $7$ & Cycle translation + B/S construction & 25.3 & 30.9 & $+3.55$ \\
    \hline
    $8$ & \verb|+GMSE| & 25.9 & 31.7 & \\
    $9$ & \verb|+reranking| & 26.3 & 32.4 & \\
    $10$ & \verb|+system combination| & 26.6 & 32.8 & \\
    \rowcolor{mypink}
    $11$ & \verb|+post processing| & \textbf{26.7} & \textbf{33.0} & $\textbf{+5.30}$\\
    \hline
    \end{tabular}
    \end{center}
    \caption{\label{tab:result}FI$\rightarrow$EN Results on newstest2018 and newstest2019. The submitted system is the last one.}
\end{table*}

\textbf{Monolingual Data}:
For our Finnish$\rightarrow$English system, back translation was performed for monolingual English data. Before back-translation, we filter them according to the aforementioned criteria in section~\ref{ssec:data} and concurrently, the scores of each sentence is obtained. After monolingual selection, there are 82M sentences remained, which is still a gigantic scale. We \textbf{\textit{cycle translate}} the last $25\%$, $50\%$ and $75\%$ of it in terms of the LM scores to empirically identify the optimal threshold and improve the fluency of monolingual corpora. In doing so, all monolingual corpus is kept at relatively high quality.\\

\begin{table}[t]
\centering
    \begin{tabular}{c|c||cc}
    \textbf{\#} &  \textbf{ $\mathcal{C}\mathcal{T}$ Ratio} & \textbf{Val.} & $\Delta$\\
    \hline\hline
    1&[$0\%$]&22.62&-\\
    2&[$25\%$]&23.18&+0.56\\
    3&[$50\%$]&\textbf{23.70}&\textbf{+1.08}\\
    4&[$75\%$]&23.07&+0.45\\
    \end{tabular}
    \caption{Different experimental settings that employed different cycle translation thresholds. Val. denotes that the results are reported on validation set.}
    \label{tab:thresholds}
\end{table}

\textbf{Synthetic Parallel Data}:
The synthetic parallel data also needs to be filtered by alignment score and word count ratio to alleviate poor translation. Further filtration retains 75M synthetic data. 

On the other hand, previous works have shown that the maximum gain can be obtained by mixing the sampled synthetic and original corpus in a ratio of 1:1~\cite{sennrich2015improving,sennrich-etal-2016-edinburgh}. The size of the synthetic corpus is generally larger than the parallel corpus, thus partial sampling is required to satisfy the 1-1 ratio. However, such sampling leads to waste of enormous synthetic data. To address this issue, we argue that a better construction strategy can be introduced to make full use of the synthetic corpus, subsequently leading to better translation quality.\\

\textbf{Small Parallel Construction}:
We randomly sampled approximate 5.8M corpus from the shuffled synthetic data for 8 times and mix them with parallel data respectively. \\

\textbf{Big Parallel Construction}:
The aim of big construction is to fully utilize the synthetic data. To achieve this, we repeated the parallel corpus 13 times and then mixed it with all synthetic corpora.

\section{Experiments}
\label{sec:exp}
The metric we employed is detokenized case-sensitive BLEU score. \verb|news-test2018| is utilized as validation set and test set is officially released \verb|news-test2019|. Training set, validation set and test set are processed consistently. Both Finnish and English sentences are performed tokenization and truecasing with Moses scripts~\cite{Koehn2007MosesOS}. In order to limit the size of vocabulary of NMT models, we adopted byte pair encoding (BPE)~\cite{sennrich-etal-2016-edinburgh} with 50k operations for each side. All the model we trained are optimized with Adam~\cite{kingma2014adam}. Larger beam size may worsen translation quality~\cite{koehn-knowles-2017-six}, thus we set beam\_size=10 for each model. All models were trained on 4 \verb|NVIDIA V100| GPUs.

In order to find the optimal threshold in cycle translation procedure, we first report our experimental results on validation data set with different thresholds, which ranges from [$0\%$, $25\%$, $50\%$, $75\%$]. Intuitively, the quality improvement of monolingual sentences afforded by cycle translation could bring better synthetic parallel data, subsequently leading to more accurate translation model. Thus, this ablation experiment was trained with synthetic parallel corpus only with different cycle translation ratios on Transformer\_base model. As is shown in Table~\ref{tab:thresholds}, when cycle translation threshold is $50\%$, the model could achieve the relatively best performance. We therefore set the cycle translation ratio to $50\%$ in our following main experiment.

Our main experiment is shown in Table~\ref{tab:result}, our baseline system is developed with the $\mathcal{M}_{small}$ configuration using the original parallel corpus and last-$20$ ensemble strategy. Unsurprisingly, the baseline system relatively performs the worst in Table~\ref{tab:result}. The $\mathcal{M}_{small}$ configuration trained with selected parallel data improves BLEU by +0.7 points. According to \textit{exp.}[3-6], adding these components can lead to continuous improvements. Notably, with Cycle Translation and Big/Small parallel construction strategy, our system could obtains +3.55 significant improvement. And \textit{exp.}[8-11] show that with performing GMSE, multi-features reranking, ConMBR system combination and post-processing, our system further improved the BLEU score from 30.9 to 33.0 on the official data set \verb|news-test2019|, which substantially outperforms the baseline by 5.3 BLEU score.

\section{Conclusion and Future Work}
\label{sec:con}
This paper presents the University of Sydney's NMT systems for WMT2019 Finnish$\rightarrow$English news translation task. We leveraged multi-dimensional strategies to improve translation quality in three levels: 1) At data level, in addition to using various data selection criteria, we proposed cycle translation to improve monolingual sentence fluency. 2) For model training, we trained multiple models with R2L corpus and big/small parallel construction corpus respectively. 3) As for inference, we prove the effectiveness of multi-features rescoring, ConMBR system combination and post-processing. We find that cycle translation and B/S construction approach bring the most significant improvement for our system.

In future work, we will apply the beam+noise method~\cite{edunov2018understanding} to generate robust synthetic data during back translation, we assume that this method combined with our proposed cycle translation strategy can bring greater improvement. Also, we would like to investigate hyper-parameter optimization for neural machine translation to avoid empirical settings.

\section*{Acknowledgments}

This research was supported by Australian Research Council Projects FL-170100117, DP-180103424 and IH-180100002. We would also thank anonymous reviewers for their comments.

\bibliography{acl2019}
\bibliographystyle{acl_natbib}
\end{document}